\documentclass[sigconf,natbib=true,anonymous=false,nonacm]{acmart}

\AtBeginDocument{%
  \providecommand\BibTeX{{%
    \normalfont B\kern-0.5em{\scshape i\kern-0.25em b}\kern-0.8em\TeX}}}

\PassOptionsToPackage{usenames,dvipsnames}{xcolor}
\usepackage{bm}
\usepackage{enumitem}
\usepackage{tikz}
\usepackage{tikz-qtree}
\usepackage{pgfplots}
\pgfplotsset{compat=1.18}
\usepackage{pgfplots}
\usepackage{subfigure}
\usepackage{adjustbox}
\usepackage{multirow}
\usepackage[normalem]{ulem}
\usepackage{soul}
\usepackage{url}
\usepackage[ruled,vlined,linesnumbered]{algorithm2e}
\usepackage{xcolor}
\usepackage{color}
\usepackage{pifont}

\usepackage{colortbl}

\usepackage{tcolorbox}
\newtcolorbox{outbox}[1]{colback=blue!5!white,colframe=blue!75!black,fonttitle=\bfseries,title=#1}

\newtcolorbox{mybox}[1]{colback=red!5!white,colframe=red!75!black,fonttitle=\bfseries,title=#1}

\begin{document}


\title{Large Language Models for Intent-Driven Session Recommendations}

\renewcommand{\shorttitle}{LLMs for Intent-Driven Session Recommendation}


\author{Zhu Sun$^{1,2,*}$, Hongyang Liu$^{3,*}$, Xinghua Qu$^{4,\dagger}$, Kaidong Feng${^5}$, Yan Wang$^3$, Yew-Soon Ong$^{1,6}$}
\affiliation{%
  \institution{$^1$Center for Frontier AI Research, A*STAR, Singapore\\
  $^2$ Institute of High Performance Computing, A*STAR, Singapore\\
  $^3$Macquarie University, Australia\\
  $^4$Shanda Group AI Lab, Singapore\\
  $^5$Yan Shan University, China\\
  $^6$Nanyang Technological University, Singapore\\
  * denotes co-first authors; $\dagger$ denotes  corresponding author: teddy.qu@shanda.com
  \country{}}
}

\renewcommand{\shortauthors}{Zhu Sun et al.}

\settopmatter{printacmref=false}

\begin{abstract}
Intent-aware session recommendation (ISR) is pivotal in discerning user intents within sessions for precise predictions. Traditional approaches, however, face limitations due to their presumption of a uniform number of intents across all sessions. This assumption overlooks the dynamic nature of user sessions, where the number and type of intentions can significantly vary. In addition, these methods typically operate in latent spaces, thus hinder the model's transparency.
Addressing these challenges, we introduce a novel ISR approach, 
utilizing the advanced reasoning capabilities of large language models (LLMs). 
{First}, this approach begins by generating an initial prompt that guides LLMs to predict the next item in a session, based on the varied intents manifested in user sessions. 
{Then}, to refine this process, we introduce an innovative prompt optimization mechanism that iteratively self-reflects and adjusts prompts. Furthermore, our prompt selection module, built upon the LLMs' broad adaptability, swiftly selects the most optimized prompts across diverse domains.
This new paradigm empowers LLMs to discern diverse user intents at a semantic level, leading to more accurate and interpretable session recommendations. Our extensive experiments on three real-world datasets demonstrate the effectiveness of our method, marking a significant advancement in ISR systems. 
\end{abstract}

\ccsdesc[500]{Information systems~Recommender systems}
\ccsdesc[300]{Computing methodologies~Neural networks}
\keywords{Session-based Recommendation,  User Intents, Large Language Models, Prompt Optimization}


\maketitle

\section{Introduction}
\begin{figure}
    \centering
    \includegraphics[width=0.9\linewidth]{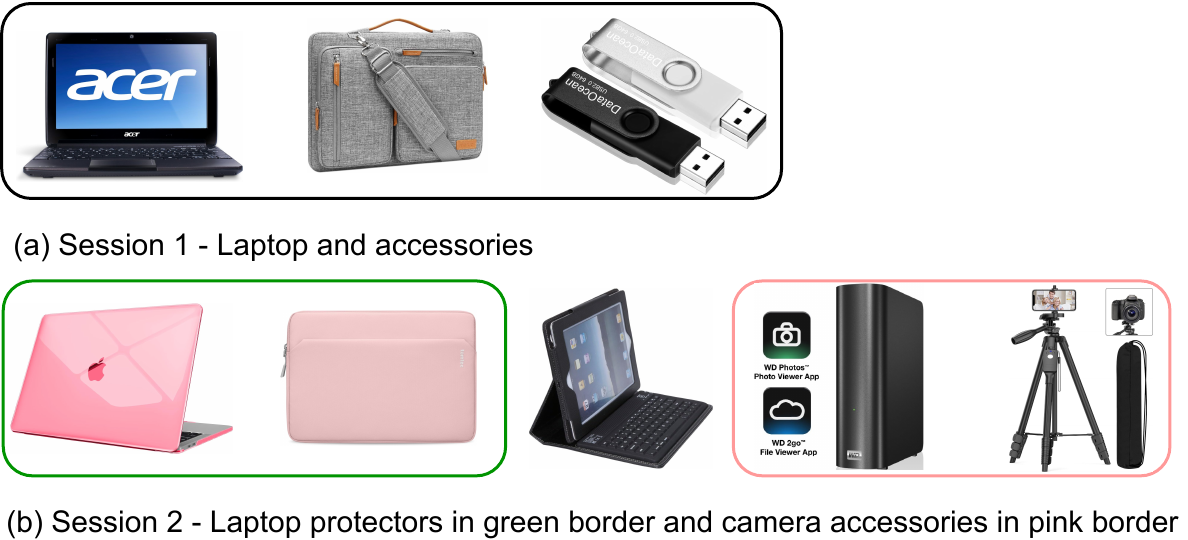}
    \vspace{-0.1in}
    \caption{Examples of user sessions with various intents.}
    \label{fig:intent-example}
    \vspace{-0.15in}
\end{figure}

Session-based recommendation (SR)~\cite{lai2022attribute,chen2022autogsr,han2022multi,hou2022core,yang2023loam} aims to predict the next interacted item based on short anonymous behavior sessions. 
Typically, different sessions may unveil diverse user intents~\cite{wang2019modeling}. Figure~\ref{fig:intent-example} illustrates two real sessions in Amazon Electronic dataset~\cite{ni2019justifying}, where the first session 
in (a) reflects one main purpose, i.e., shopping for laptop and accessories, whereas the second one 
in (b) is associated with two major intents, i.e., shopping for laptop protectors and camera accessories, respectively. 
However, most public datasets do not include explicit intents of a session as it may be intrusive and disruptive to ask users about their current session purposes directly~\cite{oh2022implicit}.
Hence, intent-aware session recommendation (ISR) has emerged to capture the latent user intents within a session, thus enhancing the accuracy of SR.

Specifically, early studies in ISR~\cite{li2017neural,liu2018stamp,wang2020global} primarily
constrain sessions to a single purpose or goal, such as shopping for a laptop and accessories in Figure~\ref{fig:intent-example}(a). However, this simplistic assumption doesn't always hold in real-world scenarios, where a session may involve diverse items
for various purposes, as illustrated in Figure~\ref{fig:intent-example}(b). It thus poses barriers for such methods seeking further performance improvements. Consequently, various approaches have been developed to model multiple intentions within a session, e.g., IDSR~\cite{chen2020improving}, MCPRN~\cite{wang2019modeling}, and NirGNN~\cite{jin2023dual}. Despite the success, they suffer from two major limitations. First, they rely on an unrealistic assumption that all sessions possess a consistent and fixed number of intentions, treating this as a hyper-parameter. Second, they are limited to learning latent intentions solely within the embedding space, greatly impeding the transparency of ISR. Such limitations thus further hinder these approaches from delivering more accurate and comprehensible recommendations.

Fortunately, the rise of large language models (LLMs) has opened up unprecedented opportunities in the field of ISR. LLMs, armed with advanced reasoning capabilities, have found widespread application in general recommendation scenario~\cite{dai2023uncovering,he2023large,sanner2023large,bao2023tallrec}, but have been relatively underexplored in the context of SR. In the limited body of research on SR, LLMs are generally employed in two distinct ways, namely in-context learning (ICL)~\cite{wang2023zero,hou2023large} and parameter-efficient fine-tuning~\cite{bao2023bi,yue2023llamaRec}. 
However, LLMs cannot fully realize their potential through simple ICL (e.g., zero-shot prompting~\cite{wang2023zero}). While fine-tuning LLMs holds promise, it grapples with challenges stemming from computational demands and the availability of open-source LLMs. 

Therefore, we propose a simple yet effective paradigm to exploit the power of LLMs for more effective ISR from the perspective of prompt optimization (abbreviated as PO4ISR). 
It is equipped with Prompt Initialization (\textit{PromptInit}), Prompt Optimization (\textit{PromptOpt}), and Prompt Selection (\textit{PromptSel}). In particular, PromptInit aims to create an initial prompt that guides LLMs to dynamically understand session-level user intents, and predict the next item accordingly. 
Inspired by the study on automatic prompt optimization~\cite{pryzant2023automatic} in natural language processing (NLP), PromptOpt
seeks to automatically optimize the initial prompt with self-reflection. To be specific, the LLM is required to offer reasoning rooted in the identified errors to improve (refine and augment) the initial prompt. The performance of improved prompts is then assessed with UCB bandits~\cite{audibert2010best}, thereby helping shortlist promising prompt candidates for an iterative optimization. 
Lastly, PromptSel prioritizes the selection of optimized prompts by 
{utilizing} the robust generalizability of LLMs across diverse domains, aiming to maximize accuracy improvements.
As such, PO4ISR can efficiently direct LLMs to infer and comprehend dynamic user intents at a semantic level, resulting in more accurate and understandable SR. 

\medskip\noindent\textbf{Contributions}. Our main contributions lie three-fold. \textbf{(1)} We introduce a simple yet powerful paradigm -- PO4ISR-- to 
utilize the capabilities of LLMs for enhanced ISR through prompt optimization. \textbf{(2)} The PO4ISR paradigm, composed of prompt initialization, optimization, and selection modules, empowers LLMs to semantically comprehend varying user intents in a session, resulting in more accurate and comprehensible SR. \textbf{(3)} Experiments on real-world datasets demonstrate that PO4ISR significantly outperforms baselines with an average improvement of 57.37\% and 61.03\% on HR and NDCG, respectively.   
Meanwhile, several insightful observations are gained, for example, 
(a) PO4ISR yields promising accuracy with only a small number of training samples; 
(b) PO4ISR exhibits advanced generalizability and excels in cross-domain scenarios;  
(c) PO4ISR showcases superior strength on sparser datasets with shorter sessions compared; however, it might exhibit increased hallucination tendencies with sparser datasets; (d) the performance of PO4ISR shows a positive correlation with the quality of the initial prompt, and lower-quality initial prompts tend to yield more significant improvements; and (f) a streamlined description and subtask division can enhance the quality of initial prompts.
\section{Related Work} 
We first provide an overview of the development of session-based recommendation (SR). Then, intent-aware session recommendation (ISR) is introduced, followed by the LLM-based SR. 

\begin{figure*}[t]
    \centering
    \includegraphics[width=0.98\textwidth]{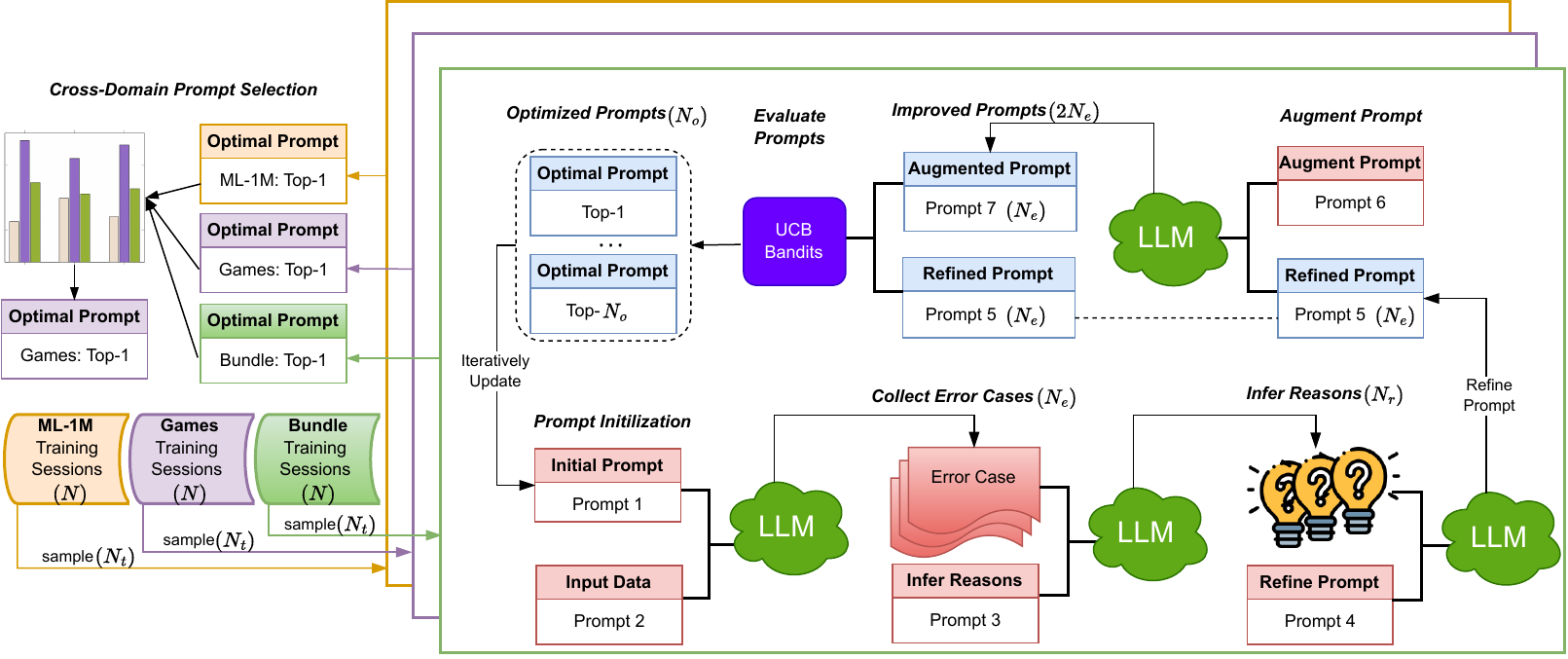}
    \vspace{-0.1in}
    \caption{The overall architecture of our proposed PO4ISR paradigm.}
    \label{fig:framework}
    \vspace{-0.1in}
\end{figure*}

\subsection{Session-based Recommendation}
Early works employ \textit{conventional methods}, such as frequent sequential patterns~\cite{ayres2002sequential,yap2012effective}, session-level item-similarity (e.g., SKNN~\cite{ludewig2018evaluation,jannach2017recurrent}), Markov chain (e.g., FPMC~\cite{rendle2010factorizing}) and random walk~\cite{choi2022s}.
Later, \textit{recurrent neural networks} (RNN) have been applied to handle longer sequences assuming that adjacent items in a session are sequentially dependent. 
GRU4Rec~\cite{hidasi2016session} is the representative model, which has been further extended 
using, e.g., data augmentation~\cite{tan2016improved,wang2022generative}, new losses~\cite{hidasi2018recurrent}, parallel architecture with item features~\cite{hidasi2016parallel}, 
and cross-session information transfer~\cite{quadrana2017personalizing}.
Subsequently, the \textit{attention mechanism} has been introduced in SR to relax this assumption by emphasizing more informative items in sessions~\cite{song2023counterfactual}, such as NARM~\cite{li2017neural} and STAMP~\cite{liu2018stamp}\footnote{Our study focuses on session-based recommendation. Therefore, some popular sequential-based recommendation approaches, e.g., Bert4Rec~\cite{sun2019bert4rec} and SASRec~\cite{kang2018self} are out of our scope. Please refer to~\cite{yin2023understanding} for the detailed difference between session- and sequential-based recommendation.}.
To better model the high-order transition among items, \textit{graph neural network} (GNN) based methods have been recently designed to generate more accurate item embeddings from the session graph, such as SR-GNN~\cite{wu2019session}, FGNN~\cite{qiu2019rethinking}, GC-SAN~\cite{xu2019graph}, GSL4Rec~\cite{wei2022gsl4rec}, 
GNG-ODE~\cite{guo2022evolutionary}, 
KMVG~\cite{chen2023knowledge},
and ADRL~\cite{chen2023attribute}.
Besides, many studies propose to 
better leverage both the intra- and inter-session information, e.g., GCE-GNN~\cite{wang2020global}, HG-GNN~\cite{pang2022heterogeneous}, 
DGNN~\cite{guo2022learning},
SPARE~\cite{peintner2023spare}, CGSR~\cite{yu2023causality}, 
and HADCG~\cite{su2023enhancing}.
Meanwhile, other works (e.g., $S^2$-DHCN~\cite{xia2021self} and CoHHN~\cite{zhang2022price}) employ hypergraphs to enhance item representations for SR. 

\subsection{Intent-aware Session Recommendation}
An essential line of research learns the intents hidden in the session for accurate SR. 
Early studies assume items inside a session are associated with one (implicit) purpose. In particular, NARM~\cite{li2017neural} extracts the last hidden state as the session representation and applies an attention mechanism on all hidden states to learn the user’s intention. 
STAMP~\cite{liu2018stamp} explicitly emphasizes the current interest reflected by the last click to capture the hybrid features of current and general interests from previous clicks via an attention network. 
SR-GNN~\cite{wu2019session}, GC-SAN~\cite{xu2019graph}, GCE-GNN~\cite{wang2020global} and TAGNN~\cite{yu2020tagnn} model each session as graph-structured data and
apply GNN to learn the representations of user intents with attention networks.
%
%
LESSR~\cite{chen2020handling} 
learns local and global interests to represent user intent by easing the information loss issue.
MSGAT~\cite{qiao2023bi} uses similar sessions to efficiently generate the session intent representation.
Nonetheless, a session may encompass items with varying intentions. Thus, solely modeling the main intent could lead to information loss, potentially hurting the performance of SR.

Consequently, recent studies endeavor to learn multiple intents for more effective SR. Specifically, 
NirGNN~\cite{jin2023dual} learns dual intents by  exploiting attention mechanisms and data distribution in the session graph. 
MCPRN~\cite{wang2019modeling} designs mixture-channel purpose routing networks to detect the purposes of each item in a session and assign them to the corresponding channels. 
IDSR~\cite{chen2020improving} projects the item representation into multiple spaces indicating various intentions and employs self-attention within each space to capture distinct intentions.
HIDE~\cite{li2022enhancing} splits an item embedding into multiple chunks to represent various intentions and then organizes items in a session with hyperedges to help learn the associated intents. 
%
MIHSG~\cite{guo2022learning} and Atten-Mixer~\cite{zhang2023efficiently} learn multi-granularity consecutive user intents to generate more accurate session representations.
STAGE~\cite{li2022spatiotemporal} and ISCON~\cite{oh2022implicit} capture the impact of multiple intrinsic intents for better SR.
DAGNN~\cite{yang2022dagnn} extracts session demands over the item category space to capture semantically correlated categories.
However, they make an impractical assumption that all sessions have a uniform and fixed number of intentions.  
Moreover, most of them can only learn latent intents, thus restricting the transparency of SR. In contrast, we aim to leverage the advanced reasoning capabilities of LLMs to uncover varying numbers of semantic intents within a session for more accurate and comprehensible SR.

\subsection{LLM-based Session Recommendation}
LLMs have achieved remarkable achievements for general recommendation~\cite{dai2023uncovering,he2023large,sanner2023large,bao2023tallrec}.
To the best of our knowledge, 
NIR~\cite{wang2023zero} is the only one that adopts zero-shot prompting for SR, and most methods target sequential recommendation~\cite{harte2023leveraging}. Among them, some leverage the in-context learning (ICL) capability of LLMs, for instance, 
\citeauthor{hou2023large}~\cite{hou2023large} use LLMs as rankers by designing sequential, recency-based and ICL prompting.
Others align LLMs for recommendation via parameter-efficient fine-tuning. To be specific,
BIGRec~\cite{bao2023bi} employ LLMs
in an all-rank scenario by grounding LLMs to the recommendation and actual item spaces.
TransRec~\cite{lin2023multi} identifies fundamental steps of LLM-based recommendation 
to bridge the item and language spaces.
GPT4Rec~\cite{li2023gpt4rec} generates multiple queries given item titles in a user's history with beam search and then retrieves items for recommendation by searching these queries with a search engine. 
LlamaRec~\cite{yue2023llamaRec} uses an ID-based sequential recommender as the retriever to generate candidates and then designs a verbalizer to transform LLM output as the probability distribution for ranking. 
Moreover,
RecInterpreter~\cite{yang2023large} examines the capacity of LLMs to decipher the representation space of sequential recommenders with sequence-recovery and -residual tasks.

However, the potential of LLMs cannot be 
utilized solely through simple ICL (e.g.,~\cite{wang2023zero}). While fine-tuning LLMs for recommendation shows promising results, it is constrained by the computational demands and availability of open-source LLMs. Instead, we introduce a new paradigm for ISR by automatically optimizing prompts, which efficiently guides LLMs to semantically comprehend the varying user intents within a session, thereby enhancing the accuracy and understandability of SR.

\section{The Proposed Method}
This section introduces the PO4ISR paradigm, a simple yet powerful framework inspired by the work~\cite{pryzant2023automatic} in the area of NLP. It is specifically designed to efficiently guide LLMs in comprehending varying user intents at a semantic level, with the goal of enhancing the accuracy and transparency of SR.

\medskip\noindent\textbf{Framework Overview}.
Figure~\ref{fig:framework} illustrates the overall framework, mainly composed of three key components. Specifically, Prompt Initialization (\textit{PromptInit}) is tasked with generating an initial prompt that directs LLMs in dynamically comprehending semantic user intents at the session level within a session. Subsequently, Prompt Optimization (\textit{PromptOpt}) aims to evaluate, refine, augment, and optimize the initial prompt through self-reflection (i.e., inferring reasons from the collected error cases). Lastly, Prompt Selection (\textit{PromptSel}) is designed to properly select optimized prompts by exploiting the robust generalizability of LLMs across diverse domains, thus maximizing the accuracy enhancements of SR.

\subsection{Prompt Initialization (PromptInit)}\label{sec:prompt-initialization}
Given a session, we first create an initial prompt for the task description. It seeks to guide LLMs in understanding the varying user intents at the semantic level, which thus empowers LLMs to make more accurate and comprehensible recommendations. The task description is demonstrated in Prompt 1 which divides the SR task into four subtasks by using the planning strategy~\cite{zhou2023least}.
Then, Prompt 1 is used to guide ChatGPT\footnote{Without a further statement, it is based on GPT-3.5-turbo.} to predict the next item based on the historical (training) user sessions fed by Prompt 2. 
For ease of understanding, we take one training session as an example, that is, the current session interactions: [1."Zenana Women's Cami Sets", 2."Monster Tattoos", 3."I Love You This Much Funny T-rex Adult T-shirt", 4."Breaking Bad Men's Logo T-Shirt", 5."Sofia the First Sofia's Transforming Dress", 6."Lewis N. Clark 2-Pack Neon Leather Luggage Tag", 7."Russell Athletic Women's Stretch Capri", 8."US Traveler New Yorker 4 Piece Luggage Set Expandable", 9."Soffe Juniors Football Capri"]. The target (ground truth) item "It's You Babe Mini Cradle, Medium" is ranked at position 19 out of 20 items in the candidate set. After using the initial prompt, the target item holds the 16th position in the re-ranked candidate set. 

\subsection{Prompt Optimization (PromptOpt)}

PromptOpt strives to evaluate, refine, augment, and optimize the initial task description prompt with an iterative self-reflection. The detailed process is elaborated in what follows.

\medskip\noindent\textbf{Collecting Error Cases}. 
We randomly sample a batch ($N_t$) of sessions from training sessions ($N$) and guide ChatGPT to predict the next items with the initial prompts.
Afterwards, we evaluate the recommendation outcomes, considering sessions where the target item ranks in the bottom half of the re-ranked candidate set as error cases. Following this rule,  the example session mentioned in Section~\ref{sec:prompt-initialization} is an error case, since the ranking position of the target item by using the initial prompt is 16 out of 20 candidates. 
These error cases, indicating that the corresponding prompts do not effectively guide LLMs in performing the SR task, will be utilized to further refine the prompts. We use $N_e$ ($0\leq N_e \leq N_t$) to denote the total number of such error cases for each batch. 

\medskip\noindent\textbf{Inferring Reasons}. Understanding the reasons behind these collected error cases would greatly aid in refining the prompts, consequently enhancing recommendation performance. Thus, we leverage the self-reflection ability of ChatGPT, that is, asking ChatGPT to reconsider and offer justifications rooted in the identified errors. Inspired by~\cite{pryzant2023automatic}, we adopt Prompt 3 to generate $N_r$ reasons for each of the $N_e$ error cases, where the `\textit{error\_case}' includes the user session and candidate item set. 
Below demonstrates the generated reasons for one error case. 
\begin{itemize}[leftmargin=*]
\item \textit{One reason why the prompt could have gotten these examples wrong is that it assumes that the user's interactive intent can be accurately inferred based solely on the items within each combination. However, the prompt does not provide any information about the user's preferences, tastes, or previous interactions. Without this context, it is difficult to accurately infer the user's intent from the items alone.}
\item \textit{Another reason is that the prompt does not specify how the combinations of items within the session should be discovered. It assumes that the combinations are already known and provided as input. However, in real-world scenarios, discovering meaningful combinations of items from a user's session interactions can be a complex task. The prompt does not provide any guidance on how to perform this discovery process, which can lead to incorrect results. }   
\end{itemize}

\medskip\noindent\textbf{Refining Prompts}. With the inferred $N_r$ reasons for each error case, we now ask ChatGPT to refine the current prompt accordingly using Prompt 4. One example of the refined prompt is illustrated as Prompt 5. 
By comparing the initial task description (Prompt 1) and the refined Prompt 5, we can easily note that the initial prompt is improved by considering two aspects: (1) user preference and taste and (2) the definition of item combinations. These two aspects are exactly consistent with the inferred reasons by ChatGPT. 

\medskip\noindent\textbf{Augmenting Prompts}. With the refined prompts, we further ask ChatGPT to augment prompts (Prompt 7 is an example of augmentation) with the same semantic meanings using Prompt 6. Accordingly, for the $N_e$ error cases, we finally obtain $2N_e$ improved prompts through refinement and augmentation. These prompts will be further utilized for an iterative optimization based on their recommendation performance, as introduced in what follows. 

\begin{mybox}{Prompt 1: \emph{Task Description}}\label{prompt:1}
\textit{Based on the user's current session interactions, you need to answer the following subtasks step by step:
\begin{itemize}[leftmargin=0.5cm]
    \item[1] Discover combinations of items within the session, where the size of combinations can be one or more.
    \item[2] Based on the items within each combination, infer the user's interactive intent for each combination.
    \item[3] Select the intent from the inferred ones that best represent the user's current preferences.
    \item[4] Based on the selected intent, please rerank the items in the candidate set according to the possibility of potential user interactions and show me your ranking results with the item index.
\end{itemize}
Note that the order of all items in the candidate set must be provided, and the items for ranking must be within the candidate set.}
\end{mybox}
\begin{mybox}{Prompt 2: \emph{Input Data}}
\textit{Current session interactions: \{[idx."item title", $\dots$]\} \\
Candidate item set: \{[idx."item title", $\dots$]\}}
\end{mybox}

\begin{mybox}{Prompt 3: \emph{Inferring Reasons for Errors}}
\textit{I'm trying to write a zero-shot recommender prompt.
\\
My current prompt is \{\textcolor{blue}{prompt}\}.
\\
But this prompt gets the following example wrong: \{\textcolor{blue}{error\_case}\}, give \{\textcolor{blue}{$N_r$}\} reasons why the prompt could have gotten this example wrong.
\\
Wrap each reason with <START> and <END>.
}
\end{mybox}
\begin{mybox}{Prompt 4: \emph{Refining Prompts with Reasons}}
\textit{I'm trying to write a zero-shot recommender prompt.
\\
My current prompt is \{\textcolor{blue}{prompt}\}.
\\
But this prompt gets the following example wrong: \{\textcolor{blue}{error\_case}\}. \\
Based on the example the problem with this prompt is that \{\textcolor{blue}{reasons}\}. \\
Based on the above information, please write one improved prompt. 
The prompt is wrapped with <START> and <END>.\\
The new prompt is:
}
\end{mybox}

\medskip\noindent\textbf{Evaluating Prompts}.
From the pool of $2N_e$ prompts, identifying the most efficient ones with the best recommendation accuracy is crucial. One greedy way is to evaluate their performance with all historical user sessions. However, this may be quite computationally expensive. To improve the efficiency, we employ the upper confidence bound (UCB) Bandits~\cite{pryzant2023automatic} to efficiently estimate the performance. In particular, it iteratively samples one prompt based on its estimated performance and then evaluates the prompt on a random batch of training sessions ($N_t$), and finally updates its performance based on the observed performance. The process is depicted by Algorithm~\ref{alg:ucb}, where the reward is calculated by the NDCG measuring the ranking position of the target item; and $\gamma$ is the exploration parameter. With UCB Bandits, we can quickly obtain the estimated performance of the $2N_e$ prompts.

\begin{outbox}{Prompt 5: \emph{The Refined Prompt}}
\textit{Given the user's current session interactions, you need to answer the following subtasks step by step:
\begin{itemize}[leftmargin=0.5cm]
    \item[1] Identify any patterns or relationships between the items within the session.
    \item[2] Based on the identified patterns, infer the user's interactive intent within each combination of items.
    \item[3] Consider the user's preferences, tastes, or previous interactions to select the intent that best represents their current preferences.
    \item[4] Rerank the items in the candidate set according to the likelihood of potential user interactions. Provide the ranking results with the item index.
\end{itemize}
Ensure that the order of all items in the candidate set is given, and the items for ranking are within the candidate set.}
\end{outbox}
\begin{mybox}{Prompt 6: \emph{Augmenting Prompts}}
\textit{Generate a variation of the following prompt while keeping the semantic meaning.
\\
Input: \{\textcolor{blue}{refined\_prompt}\}. \\
Output:
}
\end{mybox}
\begin{outbox}{Prompt 7: \emph{The Augmented Prompt}}
\textit{Please follow these steps to answer the subtasks based on the user's current session interactions:
\begin{itemize}[leftmargin=0.5cm]
    \item[1] Analyze the session items to find any patterns or relationships.
    \item[2] Use the identified patterns to determine the user's interactive intent for each combination of items.
    \item[3] Take into account the user's preferences, tastes, or previous interactions to choose the intent that best represents their current preferences.
    \item[4] Rank the items in the candidate set according to the likelihood of potential user interactions. Provide the ranking results along with the item index.
\end{itemize}
Make sure to include all items in the candidate set and only rank items within the candidate set.
}
\end{outbox}

\medskip\noindent\textbf{Iterative Optimization}. 
Based on UCB Bandits, we then iteratively optimize the prompts~\cite{pryzant2023automatic}. According to the estimated performance $R$, we select the Top-$N_o$ prompts and carry these promising prompts forward to the subsequent iteration. Specifically, we first use the Top-$N_o$ prompts to replace the prompts in the previous iteration, and then conduct the series of tasks, i.e., collecting error cases, inferring reasons, refining, augmenting, and evaluating prompts, as described in Algorithm~\ref{alg:beam-search}. 
This iterative loop fosters incremental enhancements and exploration among various promising prompt candidates. Prompt 8 is an example of an optimized prompt with the best-estimated performance in the final iteration. After applying it on the example session mentioned in Section~\ref{sec:prompt-initialization}, the target item has been re-ranked to the 11th position, advancing eight positions from its previous placement in the candidate set.   

\begin{algorithm}[t]
\footnotesize
\caption{\textsc{UCB-Bandits}}\label{alg:ucb} 
\LinesNumbered 
\KwIn{prompt set $\mathcal{P}$, training session set $\mathcal{S}$, sampled session size $N_t$, maximum epoch $E_1$, reward function $f(\cdot)$;}
\KwOut{the estimated performance $R[\mathcal{P}]$;}
    \tcp{Initialization}
    \For{each $p_i\in \mathcal{P}$}{
        $R[p_i]= 0$ \tcp*{initial estimated performance of $p_i$}
        $S[p_i] = 0$ \tcp*{initial frequency of $p_i$ being evaluated}
    }
    \For{$e_1=1; e_1 \leq E_1; e_1++$}
    {
        Sample $p_i \leftarrow \arg\; \max_p \left(R[p] + \gamma\sqrt{\frac{log(e_1)}{S[p]}}\right)$ \;
        Randomly sample $\mathcal{S}_t\subset\mathcal{S}$ where $\vert\mathcal{S}_t\vert=N_t$\;
        $r[p_i] = 0$ \tcp*{the initial accumulated reward}
        \For{each $s \in \mathcal{S}_t$}{
            $r[p_i] \leftarrow r[p_i] + f(p_i, s)$\tcp*{evaluate $p_i$ with $f(\cdot)$}
        }
        \tcp{Update evaluation frequency and performance}
        $S[p_i] \leftarrow S[p_i] + N_t$\;
        $R[p_i] \leftarrow R[p_i]+ \frac{r[p_i]}{N_t}$\;
    }
    return $R[\mathcal{P}]$\;
\end{algorithm}
\begin{outbox}{Prompt 8: \emph{The Optimized Prompt}}
\textit{Please follow these steps to answer the given subtasks:
\begin{itemize}[leftmargin=0.5cm]
    \item[1] Analyze the combinations of items in the user's session, considering any patterns or criteria.
    \item[2] Deduce the user's interactive intent within each combination, taking into account their previous interactions and preferences.
    \item[3] Determine the most representative intent from the inferred ones that aligns with the user's current preferences.
    \item[4] Reorder the items in the candidate set based on the selected intent, considering potential user interactions. Please provide the ranking results with item index.
\end{itemize}
Remember to provide the order of all items in the candidate set and ensure that the items for ranking are within the candidate set. Take into consideration the relevance of the items in the current session interactions to the candidate set, and incorporate the user's preferences and history into the recommendations.
}
\end{outbox}

\subsection{Prompt Selection (PromptSel)}
At the end of the iterative optimization, we have the Top-$N_o$ prompts. Accordingly, one straightforward way is to choose the Top-1 prompt as the final selection due to its superior overall performance. However, we notice that although a majority of sessions achieve their peak accuracy with the Top-1 prompt, a subset of them displays the best performance with other Top prompts. This can be verified by Figure~\ref{fig:same-domain-prompt}, which shows the performance gap between the Top-1 and Top-2 prompts on the validation sessions across three real-world datasets in the domains of movie, games, and e-commerce (with bundled products)\footnote{The details of the datasets utilized in our study are introduced in Section 4.1.1.}. 
From the results, it's evident that there are data points positioned where the gap is less than zero, signifying that the Top-2 prompt surpasses the Top-1 prompt in some cases. 
Thus, one can come up with potential solutions: (1) ensemble all top prompts to get the compensation results; and (2) train a classifier to select the best top prompt for each session. 
Nevertheless, we empirically discovered that (1) fails to yield promising results because the enhancements in the subset do not offset the declines in the majority; while (2) heavily relies on the accuracy of the classifier, thus bringing in extra uncertainty to the final performance. 

\begin{algorithm}[t]
\footnotesize
\caption{\textsc{Iterative-Optimization}}\label{alg:beam-search} 
\LinesNumbered 
\KwIn{$\mathcal{S}, N_t, N_o, E_{1}, E_{2}, f(\cdot)$;}
\KwOut{the optimized prompt set $\mathcal{P}_o$;}
    $\mathcal{E}\leftarrow \emptyset$, $\mathcal{P}_o \leftarrow p$\_init\; 
    \For{$e_2=1; e_2 \leq E_2; e_2++$}
    {
        $\mathcal{\tilde{P}} \leftarrow \emptyset$\;
        \For{each $p_i \in \mathcal{P}_o$}{
            Randomly sample $\mathcal{S}_t\subset\mathcal{S}$ where $\vert\mathcal{S}_t\vert=N_t$\;
            \For{each $s \in \mathcal{S}_t$}{
                \If{$s$ is an error case}{
                    $\mathcal{E} \leftarrow  \mathcal{E}.append(s)$\tcp*{collect error cases}
                }
            }
            \For{each $s \in \mathcal{E}, \vert\mathcal{E}\vert=N_e$}{
                Generate $N_r$ reasons\tcp*{infer reasons}
                $p_{i_r} \leftarrow p_{i} \& N_r$ reasons\tcp*{refine prompt}
                $p_{i_a} \leftarrow p_{i_r}$\tcp*{augment prompt}
                $\tilde{\mathcal{P}}.append(p_{i_r}, p_{i_a})$\;
            }
        }
        $R[\tilde{\mathcal{P}}] \leftarrow$ \textsc{UCB-Bandits}($\mathcal{\tilde{P}},\mathcal{S}, N_t, E_1, f(\cdot)$)\tcp*{evaluate prompts}
        $\mathcal{P}_o \leftarrow$ Top-$N_o$ of $\tilde{\mathcal{P}}$ based on $R[\tilde{\mathcal{P}}]$\tcp*{update prompts}
    }
    return $\mathcal{P}_o$\;
\end{algorithm}
\usetikzlibrary{arrows.meta,decorations.pathmorphing,backgrounds,positioning,fit,petri}




\begin{figure}[t]
    \centering
    \begin{tikzpicture}[scale=0.6]
    \pgfplotsset{%
    width=0.52\textwidth,
    height=0.38\textwidth
    }
    \begin{axis}[
        xtick={1, 25, 50, 75, 100},
        ylabel={Performance Gap w.r.t. NDCG@5},
        xlabel={Session Index},
        xlabel style ={font = \large},
        ylabel style ={font = \large},
        tick label style={/pgf/number format/fixed},
        legend style={at={(1.3,0.5)}, anchor=east,legend columns=1, row sep=0.5cm, draw=none, font=\large},
        ]
        \addplot+[only marks, mark=diamond] table [x=x, y=y, col sep=comma] {ml-1m.csv};
        \addplot+[only marks, mark=o] table [x=x, y=y, col sep=comma] {amz.csv};
        \addplot+[only marks, mark=triangle] table [x=x, y=y, col sep=comma] {games.csv};
        \legend{ML-1M, Games, Bundle}
    \end{axis}
    \end{tikzpicture}
    \vspace{-0.1in}
    \caption{Performance of Top-1\&-2 prompts (same domain).}
    \label{fig:same-domain-prompt}
\end{figure}
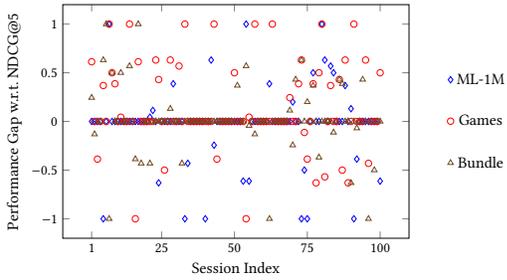

Fortunately, the robust generalizability of LLMs inspires us to explore the cross-domain performance of these optimized prompts. Specifically, Figure~\ref{fig:cross-domain-validation} depicts the performance of the Top-1 prompts from the three domains across the three datasets, for instance, `Opt-Games' (in purple) refers to the Top-1 prompt in the games domain. From the figure, we can easily notice that Opt-Games consistently performs the best not only in its games domain but also in the other two domains. One possible explanation could be attributed to the Games dataset having the shortest average session length (refer to the `Avg. Session Length' in Table~\ref{tab:statistics}) alongside a moderate level of sparsity (see `Density Indicator' in Table~\ref{tab:statistics}). These factors collectively alleviate the challenge of identifying the optimal prompt to capture crucial and unique information, thereby enhancing the overall performance of SR.
Accordingly, we select the Opt-Games as the final prompt for all domains, the efficacy of which is verified in Section 4.2.  

\begin{table}[t]
\small
\addtolength{\tabcolsep}{-1pt}
\caption{Statistics of datasets. `Density Indicator' refers to the average frequency of each item appearing in the dataset, calculated as (\#Sessions $\times$ Avg. Session Length)/\#Items.}
\label{tab:statistics}
\vspace{-0.15in}
\begin{tabular}{l|cccc}
\toprule
&\#Items &\#Sessions &Avg. Session Length &{Density Indicator} \\\midrule
ML-1M&3,416&784,860&6.85 &1573.86 \\
Games&17,389&100,018&4.18 &24.04 \\
Bundle&14,240&2,376&6.73 &1.12 \\
\bottomrule
\end{tabular}
\vspace{-0.15in}
\end{table}

\definecolor{airforceblue}{rgb}{0.36, 0.54, 0.66}
\definecolor{aliceblue}{rgb}{0.94, 0.97, 1.0}
\definecolor{alizarin}{rgb}{0.82, 0.1, 0.26}
\definecolor{almond}{rgb}{0.94, 0.87, 0.8}
\definecolor{amber}{rgb}{1.0, 0.75, 0.0}
\definecolor{amber(sae/ece)}{rgb}{1.0, 0.49, 0.0}
\definecolor{amethyst}{rgb}{0.6, 0.4, 0.8}
\definecolor{antiquebrass}{rgb}{0.8, 0.58, 0.46}
\definecolor{antiquefuchsia}{rgb}{0.57, 0.36, 0.51}
\definecolor{applegreen}{rgb}{0.55, 0.71, 0.0}
\definecolor{apricot}{rgb}{0.98, 0.81, 0.69}
\definecolor{arylideyellow}{rgb}{0.91, 0.84, 0.42}
\definecolor{ashgrey}{rgb}{0.7, 0.75, 0.71}
\definecolor{atomictangerine}{rgb}{1.0, 0.6, 0.4}
\definecolor{aureolin}{rgb}{0.99, 0.93, 0.0}
\definecolor{azure(colorwheel)}{rgb}{0.0, 0.5, 1.0}
\definecolor{babypink}{rgb}{0.96, 0.76, 0.76}
\definecolor{bluebell}{rgb}{0.64, 0.64, 0.82}
\definecolor{brightlavender}{rgb}{0.75, 0.58, 0.89}
\begin{figure}[t]
\centering
\hspace{-0.05in}
\subfigure{
\begin{tikzpicture}[scale=0.4]
\pgfplotsset{%
    width=0.46\textwidth,
    height=0.45\textwidth
}
\begin{axis}[
    ybar,
    bar width=13pt,
    ylabel={NDCG@1},
    ylabel style ={font = \Huge},
    xlabel style ={font = \Huge},
    enlarge x limits={abs=1.0cm},
    scaled ticks=false,
    tick label style={/pgf/number format/fixed, font=\Huge},
    ymin=0, ymax=0.221,
    symbolic x coords={ML-1M, Games, Bundle},
    xtick=data,
    ytick={0, 0.1, 0.2},
    legend style={at={(0.5,0.98)}, anchor=north,legend columns=4, column sep=0.2cm, draw=none, font=\Huge},
]
\addplot [fill=almond] coordinates {
(ML-1M, 0.1200) (Games, 0.1700) (Bundle, 0.1600)
};
\addplot [fill=amethyst] coordinates {
(ML-1M, 0.2100) (Games, 0.1800) (Bundle, 0.2100)
};
\addplot [fill=applegreen] coordinates {
(ML-1M, 0.1800) (Games, 0.1600) (Bundle, 0.1700)
};
\end{axis}
\end{tikzpicture}
}
\hspace{-0.1in}
\subfigure{
\begin{tikzpicture}[scale=0.4]
\pgfplotsset{%
    width=0.46\textwidth,
    height=0.45\textwidth
}
\begin{axis}[
    ybar,
    bar width=13pt,
    ylabel={NDCG@5},
    ylabel style ={font = \Huge},
    xlabel style ={font = \Huge},
    enlarge x limits={abs=1.0cm},
    scaled ticks=false,
    tick label style={/pgf/number format/fixed, font=\Huge},
    ymin=0.15, ymax=0.35,
    symbolic x coords={ML-1M, Games, Bundle},
    xtick=data,
    ytick={0.1, 0.2, 0.3},
    legend style={at={(1.56,0.5)}, anchor=east,legend columns=1, row sep=0.5cm, draw=none, font=\Huge},
]
\addplot [fill=almond] coordinates {
(ML-1M, 0.2131) (Games, 0.2492) (Bundle, 0.2205)
};
\addplot [fill=amethyst] coordinates {
(ML-1M, 0.3386) (Games, 0.3108) (Bundle, 0.3318)
};
\addplot [fill=applegreen] coordinates {
(ML-1M, 0.2734) (Games, 0.2558) (Bundle, 0.2637)
};
\legend{Opt-ML-1M, Opt-Games, Opt-Bundle}
\end{axis}
\end{tikzpicture}
}
\vspace{-0.2in}
\caption{Performance of Top-1 prompt (cross-domain).
}\label{fig:cross-domain-validation}
\vspace{-0.1in}
\end{figure}
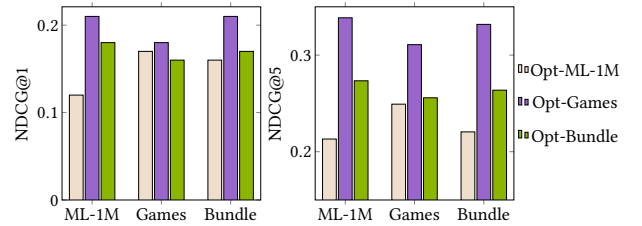

\section{Experiments and Results}

\begin{table*}[!ht]
\small
\caption{Performance comparison on all datasets, where the best and runner-up results are highlighted in bold and marked by `*'; `-' means a very small value; and `Improve' indicates the relative improvements comparing the best and runner-up results.}
\label{tab:comparison}
\addtolength{\tabcolsep}{-1pt}
\vspace{-0.15in}
\begin{tabular}{c|l|ccc|ccc|ccc|cc|cl}
\toprule
\multirow{2}{*}{Data} &\multirow{2}{*}{Metrics} &\multicolumn{3}{c|}{Conventional Methods} &\multicolumn{3}{c|}{Single-Intent Methods} &\multicolumn{3}{c|}{Multi-Intent Methods} &\multicolumn{2}{c|}{LLM Methods} &\multirow{2}{*}{Improve} &\multirow{2}{*}{$p$-value}\\\cline{3-13}
& &MostPop &SKNN &FPMC &NARM &STAMP &GCE-GNN &MCPRN &HIDE &Atten-Mixer &NIR &PO4ISR & &\\\midrule
\multirow{4}{*}{\rotatebox[origin=c]{90}{ML-1M}} 
&HR@1
&0.0004 &0.1270 &0.1132
&0.1692* &0.1584 &0.1312
&0.1434 &0.1498 &0.1490
&0.0572 &\textbf{0.2000} &18.20\% &$4.3e^{-3}$\\
&HR@5
&0.0070 &0.3600 &0.3748
&0.5230* &0.5078 &0.4748
&0.4788 &0.4998 &0.4932
&0.2326 &\textbf{0.5510} &5.35\% &$5.9e^{-2}$\\
&NDCG@1
&0.0004 &0.1270 &0.1132
&0.1692* &0.1584 &0.1312
&0.1434 &0.1498 &0.1490
&0.0572 &\textbf{0.2000} &18.20\% &$4.3e^{-3}$\\
&NDCG@5
&0.0053 &0.2530 &0.2464
&0.3501* &0.3367 &0.3044
&0.3157 &0.3256 &0.3216
&0.1436 &\textbf{0.3810} &8.83\% &$3.5e^{-3}$\\\hline
\multirow{4}{*}{\rotatebox[origin=c]{90}{Games}} 
&HR@1
&-- &0.0020 &0.0498
&0.0572 &0.0556 &0.0692
&0.0522 &0.0696 &0.0530
&0.1168* &\textbf{0.2588} &121.58\% &$6.4 e^{-5}$\\
&HR@5
&-- &0.0020 &0.2564
&0.2574 &0.2586 &0.2744
&0.2416 &0.2694 &0.2472
&0.3406* &\textbf{0.5866} &72.23\% &$3.5 e^{-5}$\\
&NDCG@1
&-- &0.0020 &0.0498
&0.0572 &0.0556 &0.0692
&0.0522 &0.0696 &0.0530
&0.1168* &\textbf{0.2588} &121.58\% &$6.4 e^{-5}$\\
&NDCG@5
&-- &0.0020 &0.1508
&0.1534 &0.1555 &0.1701
&0.1432 &0.1662 &0.1475
&0.2310* &\textbf{0.4313} &86.71\% &$7.3 e^{-6}$\\\hline
\multirow{4}{*}{\rotatebox[origin=c]{90}{Bundle}} 
&HR@1
&-- &-- &0.0398 
&0.0322 &0.0365 &0.0360 
&0.0360 &0.0458 &0.0525
&0.0975* &\textbf{0.1697} &74.05\% &$2.0 e^{-5}$\\
&HR@5
&0.0042 &-- &0.2475 
&0.2322 &0.2352 &0.2237
&0.2352 &0.2585 &0.2644
&0.2832* &\textbf{0.4328} &52.82\% &$2.6 e^{-4}$\\
&NDCG@1
&-- &-- &0.0398
&0.0322 &0.0365 &0.0360
&0.0360 &0.0458 &0.0525
&0.0975* &\textbf{0.1697}  &74.05\% &$2.0 e^{-5}$\\
&NDCG@5
&0.0021 &-- &0.1395
&0.1303 &0.1339 &0.1267
&0.1490 &0.1495 &0.1549
&0.1939* &\textbf{0.3040} &56.78\% &$3.4 e^{-5}$\\
\bottomrule
\end{tabular}
\vspace{-0.1in}
\end{table*}

We conduct extensive experiments to answer five research questions\footnote{Our code and data are available at \url{https://github.com/llm4sr/PO4ISR}}. (\textbf{RQ1}) Does PO4ISR outperform baselines? (\textbf{RQ2}) How do different components affect PO4ISR? (\textbf{RQ3}) How do essential parameters affect PO4ISR? (\textbf{RQ4}) How does PO4ISR provide comprehensible SR? (\textbf{RQ5}) Is there any limitation of PO4ISR?

\subsection{Experimental Setup}
\subsubsection{Datasets}
We use three real-world datasets from various domains. In particular, MovieLen-1M (ML-1M)\footnote{\url{https://grouplens.org/datasets/movielens/}} contains users' ratings of movies. Games is one subcategory from the Amazon dataset~\cite{ni2019justifying}, containing users' ratings towards various video games. Bundle~\cite{sun2022revisiting} contains session data for three subcategories (Electronic, Clothing, and Food) of Amazon, where the intents for each session are explicitly annotated via crowdsourcing workers. Table~\ref{tab:statistics} shows the statistics of the datasets. For ML-1M and Games, we chronologically order the rated items into a sequence for each user and then divide it into sessions by day. For each dataset, we split the sessions into training, validation, and test sets with a ratio of 8:1:1, i.e., 80\% of the initial sessions are treated as training sets; the subsequent 10\% as the validation set; and the final 10\% as the test set.  

\subsubsection{Baselines} We compare our proposed PO4ISR with ten baselines, which can be classified into three types. \textit{The first type is the conventional methods}. \textbf{Mostpop} recommends the most popular items; \textbf{SKNN}~\cite{jannach2017recurrent} recommends session-level similar items; and
\textbf{FPMC}~\cite{rendle2010factorizing} is the matrix factorization method with the first-order Markov chain.
\textit{The second type is the deep learning-based methods, which can be categorized into single-intent and multi-intent based ones}.
\textbf{NARM}~\cite{li2017neural} is an RNN-based model with the attention mechanism to capture the main purpose from the hidden states;
\textbf{STAMP}~\cite{liu2018stamp} learns the general intention of users by emphasizing the effect of the last item in the context;
\textbf{GCN-GNN}~\cite{wang2020global} uses both local and global graphs to learn item representation thus obtaining the main intent of the session; 
\textbf{MCPRN}~\cite{wang2019modeling} models users’ multiple purposes to get the final session representation; 
\textbf{HIDE}~\cite{li2022enhancing} splits the item embedding into multiple chunks, with each chunk representing a specific intention to learn diverse intentions within context;
and \textbf{Atten-Mixer}~\cite{zhang2023efficiently}: learns multi-granularity consecutive user intents to generate more accurate session representations.
\textit{The last type is the LLM-based method}.
\textbf{NIR}~\cite{wang2023zero} adopts zero-shot prompting for the next item recommendation.

\subsubsection{Parameter Settings}
We use Optuna (\url{optuna.org}) to automatically find out the optimal hyperparameters of all methods with 50 trails. In particular, the maximum training epoch is set as 100 with the early stop mechanism. 
The search space for batch size, item embedding size,
and learning rate are 
$\{64,128,256\}, \{32,64,128\}$ and $\{10^{-4},10^{-3},10^{-2}\}$, respectively.    
For SKNN, $K$ is searched from $\{50,100,150\}$. 
For NARM, the hidden size and layers are searched in $[50, 200]$ stepped by 50 and in $\{1,2,3\}$, respectively.
For GCE-GNN, the number of hops, and the dropout rate for global and local aggregators are respectively searched in $\{1,2\}, [0, 0.8]$ stepped by 0.2, and $\{0, 0.5\}$.  
For MCPRN, $\tau$ and the number of purpose channels are separately searched in $\{0.01, 0.1, 1, 10\}$ and $\{1,2,3,4\}$.
For HIDE, the number of factors is searched in $\{1,3,5,7,9\}$; the regularization and balance weights are searched in $\{10^{-5},10^{-4},10^{-3},10^{-2}\}$; the window size is searched in $[1,10]$ stepped by 1; and the sparsity coefficient is set as 0.4.
For Atten-Mixer, the intent level $L$ and the number of attention heads are respectively searched in $[1,10]$ stepped by 1 and in $\{1,2,4,8\}$.
For PO4ISR, $N=50, N_t=32, N_r=2, N_o=4, E_1=16, E_2=2$; and we randomly select 8 prompts as the input of the UCB Bandits in each iteration.  

\subsubsection{Evaluation Metrics}
Following state-of-the-arts~\cite{yin2023understanding,wang2019modeling,wang2023zero}, HR@K and NDCG@K are adopted as the evaluation metrics. We set $K=1/5$ since most users tend to prioritize the quality of items appearing at the top positions in real scenarios~\cite{liu2021top}. Generally, higher metric values indicate better ranking results. 
For a fair comparison, the candidate size of all methods is set as 20 following NIR~\cite{wang2023zero};
all non-LLM baselines are trained with 150 randomly sampled sessions from the training set, while PO4ISR uses the subset of 50 sessions. This is because although PO4ISR uses 50 sessions to optimize prompts in each domain, it considers the performance from (three) diverse domains for the final prompt selection. For non-LLM baselines, it is non-trivial to use the cross-domain performance to select the best models. This also helps showcase the superior advantages of LLMs on generalization.  
To manage the API call cost, for ML-1M and Games, we randomly sample 1000 sessions from the validation set and test set, respectively. 
For robust performance, we repeat the test procedure five times where each time we set different seed values (i.e., 0, 10, 42, 625, and 2023) to generate different candidate sets. Finally,  we report the average results as presented in Table~\ref{tab:comparison}.  

\begin{table*}[t]
\small
\caption{The results of ablation study across all datasets on the test set (seed 0).}\label{tab:ablation}
\addtolength{\tabcolsep}{-1.5pt}
\vspace{-0.15in}
\begin{tabular}{l|ccccc|cccc|ccccc}
\toprule
&\multicolumn{5}{c|}{ML-1M} 
&\multicolumn{4}{c|}{Games}
&\multicolumn{5}{c}{Bundle}\\
&Initial &Top-1 &EnSame &EnCross &PO4ISR   
&Initial &EnSame &EnCross &PO4ISR 
&Initial &Top-1 &EnSame &EnCross &PO4ISR \\\midrule
HR@1 
&0.1430 &0.2070 &0.1120 &0.1070 &\textbf{0.2110}
&0.0790 &0.1540 &0.1090 &\textbf{0.2600}
&0.0504 &0.1176&0.0294 &0.0840 &\textbf{0.1933}\\
HR@5 
&0.4150 &0.5130&0.4130 &0.4250 &\textbf{0.5730}
&0.3510 &0.5250 &0.4410 &\textbf{0.5960}
&0.2437&0.3193&0.1891 &0.2689 &\textbf{0.4454}\\
NDCG@1 
&0.1430 &0.2070&0.1120 &0.1070 &\textbf{0.2110}
&0.0790 &0.1540 &0.1090 &\textbf{0.2600}
&0.0504&0.1176&0.0294 &0.0840 &\textbf{0.1933}\\
NDCG@5 
&0.2823 &0.3662&0.2693 &0.2640 &\textbf{0.3975}
&0.2110 &0.3574  &0.2779 &\textbf{0.4381}
&0.1396&0.2202&0.1119 &0.1745 &\textbf{0.3183}\\
\bottomrule
\end{tabular}
\vspace{-0.1in}
\end{table*}

\input{plot/cross-domain-test}

\subsection{Results and Analysis}

\subsubsection{Overall Comparison (RQ1)}

Table~\ref{tab:comparison} shows the performance of all methods on the three datasets. Several findings are noted. 

\textbf{(1)} Regarding conventional methods (CMs), the model-based FPMC performs the best on sparser datasets Games and Bundles, while is slightly defeated by SKNN on the denser dataset ML-1M (see `Density Indicator' in Table~\ref{tab:statistics}). This exhibits the strong capability of model-based methods in learning sequential patterns with sparse data. \textbf{(2)} For single-intent methods (SIMs), each gains its best performance on different datasets, e.g., NARM excels on ML-1M, whereas GCE-GNN wins on Games. Generally, GCE-GNN showcases advantages on shorter sessions with sparser datasets (e.g., Games). This is because such sessions do not contain sufficient local information, thus requiring global information (graph) to compensate. Conversely, NARM performs well on longer sessions with denser datasets (e.g., ML-1M), since such sessions possess substantial information (e.g., frequent patterns) making the fusion of global information potentially noisy. \textbf{(3)} Concerning multi-intent methods (MIMs), MCPRN displays the poorest performance across all datasets, as limited training data hinders the learning of proper parameters for multiple RNN channels. Besides, HIDE exceeds Atten-Mixer on relatively denser datasets (ML-1M and Games) but lags on the sparsest Bundle.
\textbf{(4)} Among LLM-based methods (LLMMs), PO4ISR outperforms NIR, corroborating our prior assertion that basic in-context learning fails to fully exploit LLMs, thereby emphasizing the superiority of our prompt optimization paradigm.  

Overall, CMs underperform SIMs or MIMs, underscoring the importance of capturing user intents for improved SR. The victory of MIMs over SIMs on Bundle validates the effectiveness of multi-intent learning. However, some of the SIMs (e.g., NARM) surpass MIMs (e.g., MCPRN) on ML-1M and Games, suggesting that fixed numbers of intents might constrain the capability of MIMs. LLMMs exhibit strength on sparser datasets compared to denser ones (e.g., NIR on Games vs. ML-1M), and shorter sessions compared to longer ones (e.g., NIR on Games vs. Bundles). Similar trends are also observed in PO4ISR, where the average improvement (12.65\%) on ML-1M  is smaller than that (100.55\%) on Games. This demonstrates the advanced ability of LLMs to address the data sparsity issue. Lastly, our PO4ISR consistently achieves the best performance among all baselines, with an average improvement of 57.37\% and 61.03\% on HR and NDCG, respectively.   

\subsubsection{Ablation Study (RQ2)}
To verify the efficacy of different components, we compare PO4ISR with its four variants. In particular, `Initial' means we only use the initial prompt without optimization for each domain; `Top-1' means we merely adopt the Top-1 prompt within each domain; `EnSame' means we ensemble the ranking results of both Top-1 and -2 prompts within each domain; `EnCross' means we ensemble the ranking results of Top-1 prompts across all domains. The performance is shown in Table~\ref{tab:ablation}. 
Three observations are noted: (1) Initial performs worse than Top-1, which indicates the effectiveness of iterative prompt optimization; (2) both EnSame and EnCross underperform Top-1, implying that simply ensembling the top prompts either within the same domain or across different domains cannot efficiently improve the performance; and (3) PO4ISR with the Top-1 prompt in the games domain consistently achieves the best performance across all datasets, showcasing the efficacy of our cross-domain prompt selection strategy. This is further confirmed by Figure~\ref{fig:cross-domain}, presenting the performance of Top-1 prompt from each domain in the cross-domain scenario.

\subsubsection{Parameter Analysis (RQ3)}
We further investigate the impact of important hyperparameters on PO4ISR. 
First, we examine the impact of different initial prompts. To this end, we substitute the initial task description (Prompt 1) with Prompt 9 incorporating two major changes: (1) we use `preferences' to replace `intentions', and the two terms have the same meanings in the context of ISR, and (2) we simplify the four subtasks into two subtasks.   
The results are presented in Table~\ref{tab:parameter-analysis} (rows 1-3 vs. rows 4-6), where we note that directly using Prompt 9 achieves better performance than using Prompt 1; meanwhile, the corresponding Top-1 prompt optimized based on Prompt 9 outperforms the Top-1 prompt optimized based on Prompt 1. This indicates that (1) simplified descriptions and subtasks division can improve the quality of the initial prompt, \textit{vice versa}; 
(2) the quality of the initial prompts positively affects the final performance; 
(3) regardless of the quality of initial prompts, they can be largely enhanced with iterative optimization; (4) the lower-quality initial prompt yields larger overall improvements;   
and (5) the performance of PO4ISR presented in Table~\ref{tab:comparison} is not the upper bound, and can be further improved with better initial prompts, showcasing its great potential. 
Besides, we study the impact of batch size $N_t$ on the final performance by varying its values in $\{16, 32\}$, and the results are presented in Table~\ref{tab:parameter-analysis} (row 2 vs. row 7). Accordingly, we find that $N_t=32$ is the optimal setting. 

\begin{mybox}{Prompt 9: \emph{Task Description}}
\textit{Based on the user's current session interactions, you need to answer the following tasks:
\begin{itemize}[leftmargin=0.5cm]
    \item[1] Please infer the user's preferences, considering that the user may have one or multiple preferences.
    \item[2] Based on inferred preferences, please rerank the items in the candidate set according to the possibility of potential user interactions and show me your ranking results with the item index.
\end{itemize}
Note that the order of all items in the candidate set must be provided, and the items for ranking must be within the candidate set.}
\end{mybox}

\subsubsection{Visualization (RQ4)} 
\begin{figure}[t]
    \centering
    \includegraphics[width=0.95\linewidth]{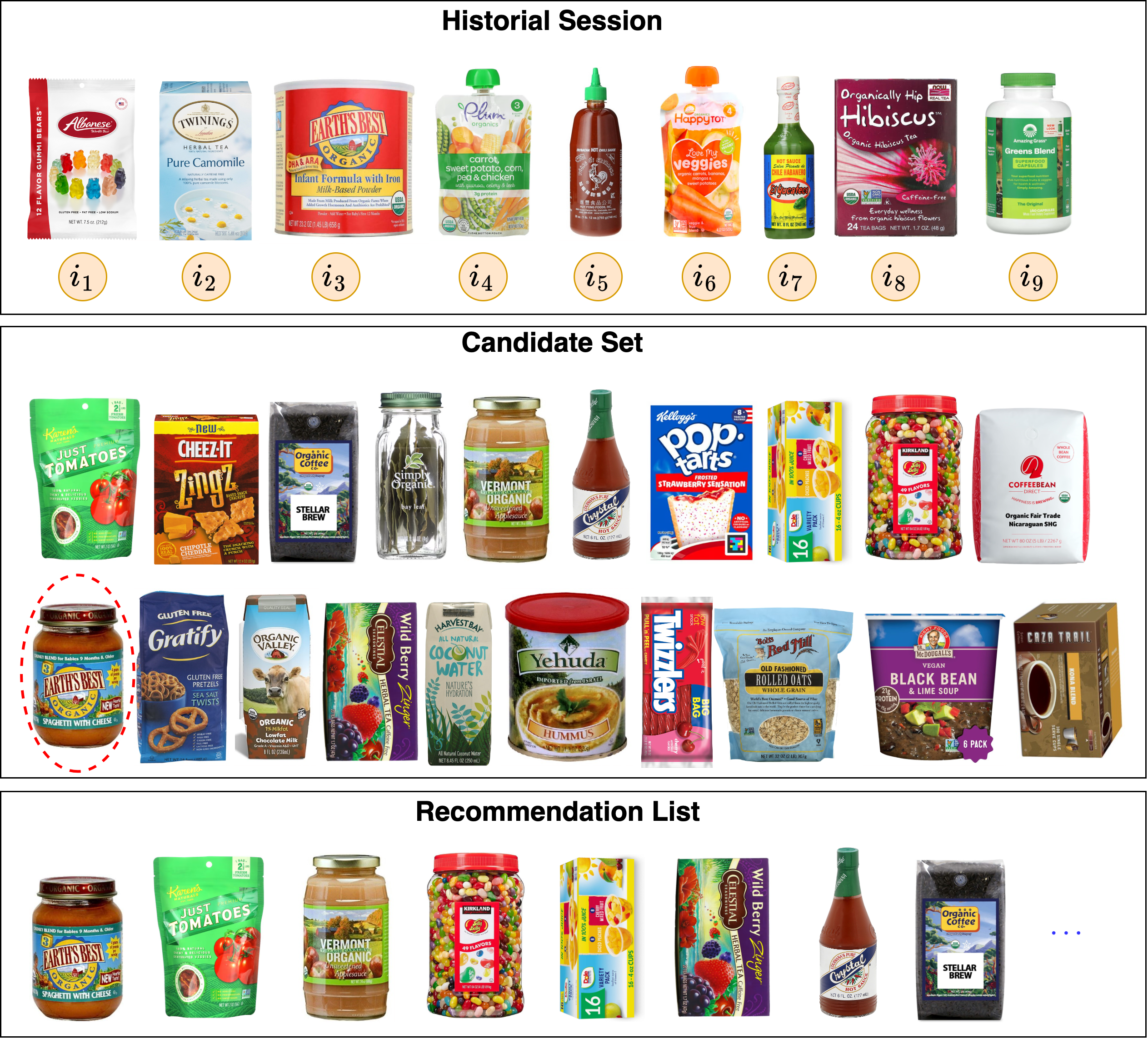}
    \vspace{-0.1in}
    \caption{The case study on Bundle.}
    \label{fig:case-study}
    \vspace{-0.1in}
\end{figure}

To illustrate the results generated by PO4ISR, we randomly sample one test session from Bundle as shown in Figure~\ref{fig:case-study}. Given the historical session, PO4ISR first detects four combinations of items, including $[i_1, i_2, i_8], [i_3, i_4, i_6], [i_5, i_7], [i_9]$. The corresponding intents for each combination are (1) "this combination includes gummi bears, chamomile tea, and rosehip with hibiscus tea; the user might be interested in snacks and beverages for relaxation or enjoyment"; (2) "this combination includes organic infant formula, organic baby food, and organic baby food pouches; the user's intent seems to be focused on organic and healthy options for infants"; (3) "this combination includes Sriracha hot sauce and green hot sauce; the user's intent appears to be related to spicy condiments or sauces" and (4) "this combination includes green superfood capsules; the user's intent might be to explore or maintain a healthy lifestyle". Then it identifies and outputs the most important intent: 
"based on the given combinations, the intent related to organic and healthy options for infants (Combination 2) seems to be the most specific and focused". Finally, it furnishes a re-ranked recommendation list, positioning the ground truth item "Earth's Best Organic Baby Food, Spaghetti With Cheese" (highlighted in a red-dot circle) from its initial 11th rank within the candidate set to the topmost position. In summary, the case study validates that PO4ISR can help provide more comprehensible recommendations. 

\begin{table}[t]
\centering
\small
\addtolength{\tabcolsep}{-1.5pt}
\caption{Results of parameter analysis on Bundle (seed 0).}
\label{tab:parameter-analysis}
\vspace{-0.15in}
\begin{tabular}{c|cccc}
\toprule
&HR@1 &HR@5 &NDCG@1 &NDCG@5  \\\midrule
Prompt 1 (Initial) &0.0504 &0.2437 &0.0504 &0.1396\\
Top-1 ($N_t=32$) &0.1176 &0.3193 &0.1176 &0.2202 \\
Improve 
&\cellcolor{blue!40}133.33\% &\cellcolor{blue!40}31.02\% &\cellcolor{blue!40}133.33\% &\cellcolor{blue!40}57.74\%\\\hline
Prompt 9 (Initial) &0.1008 &0.2479 &0.1008 &0.1706\\
Top-1 ($N_t=32$) &0.1807 &0.3739 &0.1807 &0.2744\\
Improve 
&\cellcolor{blue!25}79.27\% &\cellcolor{blue!25}50.83\% &\cellcolor{blue!25}79.27\% &\cellcolor{blue!25}60.84\%\\\hline
Top-1 ($N_t=16$) &0.1008 &0.2857 &0.1008 &0.1860\\
Improve (vs. Prompt 1) &\cellcolor{blue!15}100\% &\cellcolor{blue!15}17.23\% &\cellcolor{blue!15}100\% &\cellcolor{blue!15}33.24\% \\
\bottomrule
\end{tabular}
\vspace{-0.1in}
\end{table}

\subsubsection{Discussion on Hallucination (RQ5)} 

Despite the success of PO4ISR on the SR task, it showcases limitations due to the inherent issue of LLMs. One issue is that PO4ISR may generate hallucination for some sessions (e.g., the response does not contain the ranking list or the ground truth item is not included in the ranking list), although we add hard constraints in the prompt such as "\textit{the order of all items in the candidate set must be provided, and the items for ranking must be within the candidate set}". Table~\ref{tab:bad-case} illustrates the ratio of sessions having hallucinations on the test sets using the Top-1 prompt from each domain across the three datasets. Two major observations can be noted. 
\textbf{(1)} The ratio is at its lowest on ML-1M but peaks on Bundle (emphasized in blue). This discrepancy likely stems from ML-1M having the highest average repeat frequency of items across sessions, whereas Bundle exhibits the lowest trend (see Table~\ref{tab:statistics}). The frequent appearance of items in various sessions may simplify the pattern recognition process, thus reducing task complexity to some extent. Additionally, Bundle's diverse range of electronics, clothing, and food products elevates the complexity of the SR task compared to the more focused ML-1M dataset. \textbf{(2)} The optimal prompts from different domains show comparable performance as highlighted in pink. On average, there are around 5.55\% sessions with hallucination, 
implying that further performance enhancements can be obtained by addressing this issue and exhibiting the latent potential of LLMs for SR (note that the HR and NDCG values are set as 0 for sessions with hallucination).            
%

To alleviate the hallucination issue, we involve JSON mode\footnote{\url{https://platform.openai.com/docs/guides/text-generation/json-mode}} in the response to better control the output by adding the constraint as: \textit{Provide the ranking results for the candidate set using JSON format, following this format without deviation: [{"Item ID": "correspond item index", "Item Title": "correspond Item Title"}]}.
Table~\ref{tab:json} illustrates the performance contrast of PO4ISR on Bundle, with and without the JSON mode in the response, which indicates that employing JSON mode marginally reduces bad cases, yet significantly compromises recommendation accuracy.
Contrarily, we find most such cases can be better eased by using GPT-4 with less compromised accuracy. 
%
\begin{table}[t]
\small
\caption{Ratio of sessions with hallucination (seed 0).}\label{tab:bad-case}
\addtolength{\tabcolsep}{1pt}
\vspace{-0.15in}
\centering
\begin{tabular}{c|ccc|c}
\toprule
&{Opt-ML-1M} &{Opt-Games} &{Opt-Bundle} &Average \\\midrule 
ML-1M &0.30\% &0.10\% &0.30\% & \cellcolor{blue!15} 0.23\% \\
Games &8.30\% &6.80\% &6.00\% &\cellcolor{blue!25}7.03\%\\
Bundle &7.56\% &9.66\%&10.92\% &\cellcolor{blue!45}9.38\%\\\hline
Average &\cellcolor{red!15}5.39\% &\cellcolor{red!15}5.52\% &\cellcolor{red!15}5.74\% &5.55\%\\
\bottomrule
\end{tabular}
\vspace{-0.1in}
\end{table}

\begin{table}[t]
\small
\caption{Performance with and without JSON on Bundle.}\label{tab:json}
\addtolength{\tabcolsep}{-0.5pt}
\vspace{-0.15in}
\centering
\begin{tabular}{l|cc|cc|cc}
\toprule
&\multicolumn{2}{c|}{Opt-ML-1M}
&\multicolumn{2}{c|}{Opt-Games}
&\multicolumn{2}{c}{Opt-Bundle} \\\cline{2-7}
&\cellcolor{blue!15}+JSON &\cellcolor{red!15}-JSON &\cellcolor{blue!15}+JSON &\cellcolor{red!15}-JSON &\cellcolor{blue!15}+JSON &\cellcolor{red!15}-JSON \\\midrule 
HR@1 &0.0546 &0.1429 &0.0504 &0.1933 &0.0756 &0.1176\\
HR@5 &0.2227 &0.3361 &0.2185 &0.4454 &0.2395 &0.3193\\
NDCG@1 &0.0546 &0.1429 &0.0504 &0.1933 &0.0756 &0.1176\\
NDCG@5 &0.1336 &0.2355 &0.1289 &0.3183 &0.1531 &0.2202\\\hline
Ratio &\cellcolor{blue!15}7.14\% &\cellcolor{red!15}7.56\% &\cellcolor{blue!15}8.82\% &\cellcolor{red!15}9.66\% &\cellcolor{blue!15}9.66\% &\cellcolor{red!15}10.92\%\\
\bottomrule
\end{tabular}
\vspace{-0.1in}
\end{table}


\section{Conclusion}
Inspired by the reasoning capability of LLMs, we introduce a new paradigm -- PO4ISR -- for intent-aware session recommendation. It aims to discover varying numbers of semantic intents hidden in different sessions for more accurate and comprehensible recommendations through iterative prompt optimization. Specifically, the Prompt Initialization module first creates the initial prompt to instruct LLMs to predict the next item by inferring varying intents reflected in a session. Then, the prompt optimization module is devised to optimize prompts with iterative self-reflection in an automatic manner. Finally, the prompt selection module seeks to appropriately select optimal prompts based on the robust generalizability of LLMs across diverse domains.  
Extensive experiments on real-world datasets show the superiority of PO4ISR against other counterparts. Furthermore, several insightful discoveries are made to guide subsequent studies in this area.

\bibliographystyle{ACM-Reference-Format}
\bibliography{reference}


\end{document}